# SINGLE FRAME IMAGE SUPER RESOLUTION USING LEARNED DIRECTIONLETS

Reji A P and Tessamma Thomas

Department of Electronics, Cochin University of Science and Technology , Cochin 22, India

reji@cusat.ac.in
tess@cusat.ac.in

## ABSTRACT

*In this paper, a new directionally adaptive, learning based, single image super resolution method using multiple direction wavelet transform, called Directionlets is presented. This method uses directionlets to effectively capture directional features and to extract edge information along different directions of a set of available high resolution images .This information is used as the training set for super resolving a low resolution input image and the Directionlet coefficients at finer scales of its high-resolution image are learned locally from this training set and the inverse Directionlet transform recovers the super-resolved high resolution image. The simulation results showed that the proposed approach outperforms standard interpolation techniques like Cubic spline interpolation as well as standard Wavelet-based learning, both visually and in terms of the mean squared error (mse) values. This method gives good result with aliased images also.*

## 1. INTRODUCTION

Most of the Image processing applications like remote sensing, medical imaging, robot vision, industrial inspection, or video enhancement demand high resolution images. The high resolution images not only give the viewer a pleasing appearance but also offer additional information that is important for the analysis in many applications. For any recognition system, blurred or noisy images are a head ache and also many segmentation algorithms do not work well when images are blurred. Acquisition environment condition, the resolution of image sensors employed, etc are some of the factors that affect the quality of digital image .Getting high quality images in practical applications like satellite imaging is a difficult task since the above factors cannot be controlled. In the case of satellite imaging, the distance between the earth and satellite cannot be reduced and weather conditions cannot be controlled. High resolution images mainly depend on sensor manufacturing technology that tries to increase the number of pixels per unit area by reducing the pixel size. But there is limitation to pixel size reduction due to shot noise encountered in the sensor itself and the high precision optics sensors are too expensive to use for commercial applications. Therefore some image processing methods are needed to construct a high resolution image from one or more available low resolution images. Super resolution refers to the process of producing a high resolution image than what is afforded by the physical sensor through post processing ,making use of one or more low resolution observations[1] .It includes up sampling the image ,thereby increasing the maximum spatial frequency ,and removing degradations that arise during the image capture ,namely ,aliasing and blurring.

Super resolved image reconstruction is proved to be effective in many areas including medical imaging, satellite imaging, video applications, image enlarging in web pages and restoration of old historic photographs ,surveillance, tracking, and license plate recognition system etc. Techniques such as bilinear and bicubic interpolation only consider low resolution image information and the resulting image from these techniques is often blurry and contain artefacts. In general there are two types of super resolution techniques- reconstruction based and learning based. In reconstruction based techniques, high resolution image is recovered from several low resolution observations of the input. Impressive amount of work has been reported in this field. Frequency domain approach proposed by Tsai and Huang [3] was the first method in super resolution. In 1990, Kim et al. proposed a recursive algorithm for restoration of super resolution images from noisy and blurred observations [7].M Irani and S Peleg   proposed an approach which   was similar to back-projection used in tomography [24]. Tekalp et al. addressed the super resolution problem as reconstruction of high resolution image from low resolution frames of the same scene where the successive frames were uniformly shifted versions of each other at sub pixel displacements [25]. The algorithm proposed by M .Elad and, Feuer was a generalization of the stochastic estimation based methods (the ML and the MAP estimators) for the restoration of single blurred and noisy images[26]. Ur and Gross used the Papoulis and Brown generalised sampling theorem for obtaining high resolution image from a set of spatially shifted observations [8]. Ng et al. developed a regularised, constrained total least squares solution to obtain a high resolution image [10]. Nguyen et al. have proposed a circulant block pre conditioners to accelerate the conjugate descent method while solving the Thikonov-regularised super resolution problem [11].Reconstruction based super resolution methods is applicable only if the images are sub-pixel shifted.

But in learning based super resolution algorithms, a training set of available high resolution images are used to obtain the high resolution of an image captured using a low resolution camera. In the training set, the images are stored as patches or as coefficients of other feature representations like Wavelet transform, DCT etc. Unlike the reconstruction based method which requires multiple low resolution input images, here only one input image (single frame image super resolution) is required. Single frame image super resolution can be used in applications where database of high resolution images are available.  These methods are classified under the motion free super resolution scheme.

In [2] Freeman proposed an example based super resolution method in which he had developed a Bayesian propagation algorithm using Markov Network. Bishop et al. [12], Pickup et al.[13], and Sun et al. [14] also adopted the Markov network as Freeman and Pasztor , did but they differed in the definition of priors and likelihoods. Baker and Kanade's hallucination algorithm [15] further inspired the work in this field. Gunturk et al. [16], Liu et al. [18], and Wang and Tang [19] all used face bases and inferred the combination coefficients of the bases, where the face bases are different. Liu et al.'s face hallucination algorithm [20] was a combination of [21] and [22] to infer the global face structure and the local details respectively. In [17]  D Capel et al explored learning of some constrained models and priors from training set images and applying these to super resolution restoration. In [27] Jian Sun et al.proposed a Bayesian approach to image hallucination.In [28]  Midhun Das Gupta et al presented a novel learning based method for restoring and recognizing images of digits that had been blurred using an unknown kernel. The basic idea of Qiang Wang et al work [29] was to bridge the gap between a set of low resolution (LR) images and the corresponding high resolution (HR) image using both the SR reconstruction constraint and a patch based image synthesis constraint in a general probabilistic framework. In [30] Kwang In Kim et al  proposed a regression-based method for single imag super-resolution. Kernel ridge regression (KRR) was used to estimate the high-frequency details of the underlying high-resolution image. Joshy and Choudhari have proposed a learning based method for image super resolution from zoomed observations. They model high resolution image as Markov random field, the parameters of which are learned from the most zoomed observation [23]. The learned parameters are then used to obtain a maximum aposteriori estimate of the high resolution image. Several learning based methods based on wavelets are also available [1, 4]. But wavelets are isotropic and cannot follow edges in images which are oriented along arbitrary directions.
This produces artefacts in the reconstructed image.
The proposed learning based method is motivated by the work [5], in which directionlets are proved to provide sparse representation of images like wavelets. This novel lattice-based transform exploits multi-directional and retains the simplicity of computations and filter design from the standard WT [5]. This multi-directionality and anisotropy overcome the weakness of the standard WT in presence of edges and contours, that is, they allow for sparser representations of these directional anisotropic features

## 2. LEARNING BASED SUPER RESOLUTION

### 2.1. Low resolution model

 Single frame super-resolution algorithms attempt to estimate high resolution image from single low resolution observation. This is considered as an inverse problem. For solving such inverse problem needs to devise a forward model that represents the image formation process. The decimation model for obtaining low resolution image y (i ,j) can be obtained from its high resolution version z(k ,l) as

$$y(i,j) = \frac{1}{q^{\wedge}2} \sum_{k=qi}^{q(i+1)-1} \sum_{l=qj}^{q(j+1)-1} z(k,l) \qquad (1)$$

The low resolution pixel intensity y(i ,j) is the average of high resolution intensities over a neighbourhood of $q^{\wedge}2$ pixels[1]. Usually super resolution problem can be modelled as,

$$Y = DBz + n \qquad (2)$$

Here y represents the lexicographically ordered vector of size $M^{\wedge 2}$ x1, which is formed from the observed low resolution image Y of size MxM. Similarly, z is the high resolution image to be super resolved. D is the decimation matrix whose size depends on the decimation factor q. For an integer decimation factor of q, the decimation matrix D consists of $q^{\wedge}2$ non-zero elements along each row at appropriate locations. Here n is the independent and identically distributed (i.i.d.) noise vector with zero mean and variance $\sigma^{\wedge 2}$. It has same size as y. Here the problem is to estimate z given y, which is an ill-posed inverse problem. It may be mentioned here that the observation captured is not blurred. In other words, it assumes identity matrix for blur. Generally, the decimation matrix to obtain the aliased pixel intensities from the high resolution,

$$D = \frac{1}{q^{\wedge}2} \begin{bmatrix} 11...1 & & & 0 \\ & 11...1 & & \\ & & . & \\ & & & . \\ 0 & & & 11...1 \end{bmatrix} \qquad (3)$$

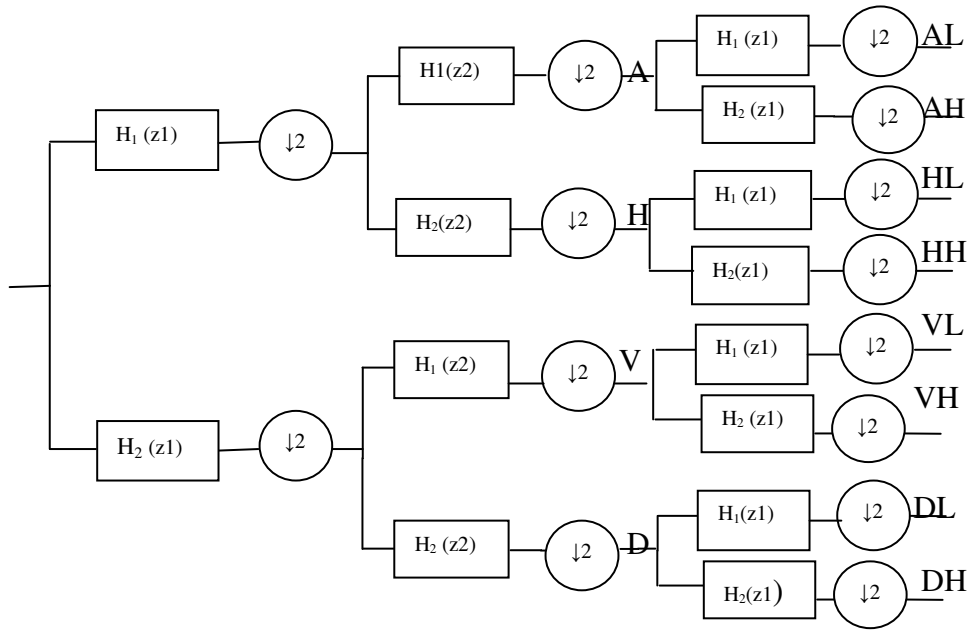

Fig 1.Filtering scheme for the AWT (2, 1), where one step of iteration is shown

For example ,the decimation factor of q=2 and with lexicographically ordered z of size ,say 16*1 ,the D matrix is of size 4*16 and can be written as

$$D = \frac{1}{4} \begin{bmatrix} 1100110000000000 \\ 0011001100000000 \\ 0000000011001100 \\ 0000000000110011 \end{bmatrix} \qquad (4)$$

The Eq. (2) indicates that a low resolution pixel intensity y(i; j) is obtained by averaging the intensities of q2 pixels to the same scene in the high resolution image and adding noise intensity n(i; j).

## 2.2. Directionlets

It is already proved that the standard wavelet transform is an efficient tool for analyzing one dimensional signal. But it is isotropic in the sense that filtering and sub sampling operations are applied equally along both horizontal and vertical directions at each scale .This isotropic transform cannot properly capture the anisotropic discontinuities present in the two dimensional signals like images. This is because the directions of the transforms and discontinuities in images are not matched and the transforms fail to provide a compact representation of two dimensional signals. That is, the problem of standard wavelet transform is that only vertical and horizontal directions are considered and number of iteration on both directions is equal. The directionlet transform is anisotropic. The difference between isotropic and anisotropic wavelet transform is that in the anisotropic wavelet transform, the number of transforms applied along the horizontal and vertical directions is unequal, that is there are n1 horizontal and n2 vertical transforms at a scale, where n1 is not necessary equal to n2.The iteration process is continued in the low sub band, like in the standard wavelet transform. Anisotropic transform is represented as AWT (n1, n2). The standard WT is simply given by AWT (1, 1).The anisotropic ratio $\rho = n1/n2$ determines elongation of the basis functions of the AWT (n1, n2).When n1=2, n2=1, the AWT (2, 1) produces eight bands AL, AH, HL, HH, VL, VH, DL and DH as in figure 1. Figure 2(a) and figure 2(b) shows frequency decomposition of 2-D WT and AWT .The skewed AWT can trace the discontinuity efficiently with fewer significant coefficients compared with wavelet as shown in figure 2(c) and figure 2(d). The directionlets are skewed anisotropic Wavelet transform(S-AWT).That means scaling and filtering operations are along a selected pair of directions, not necessarily horizontal and vertical. The 1-D filtering and sub sampling operations of standard wavelet transform is retained here and can provide anisotropic perfect reconstruction, unlike in the case of some other directional transform constructions (e.g. curvelets, contourlets or edgelets).

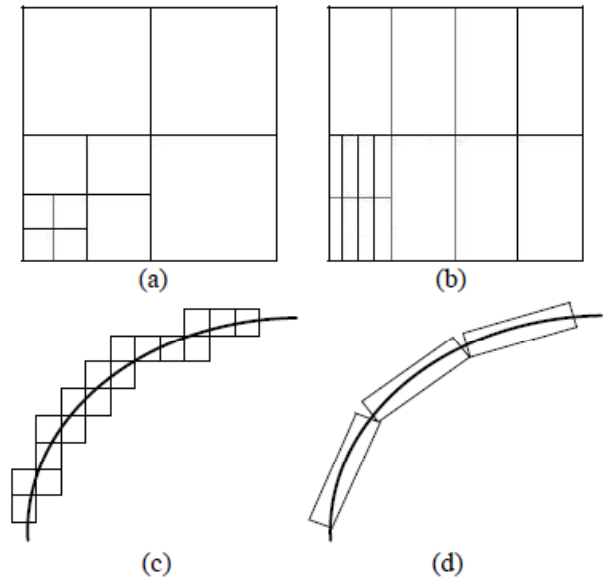

Figure.2 Frequency decomposition of (a) 2-D WT (b) AWT (c) isotropic basis function (d) multi directional (skewed) anisotropic basis function[10].

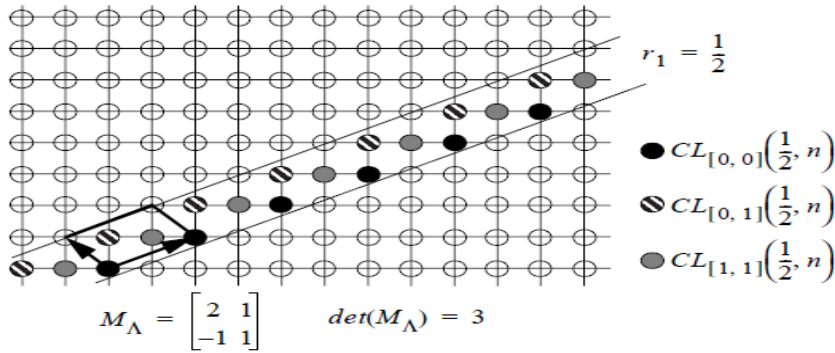

Figure 3 The intersections between the 3 cosets of the lattice Λ given by the generator matrix M and the digital lines L(r1 = 1/2, n), where n ∈     , are the co-lines CL[0,0](1/2, n), CL[0,1](1/2, n), and CL[1,1](1/2, n)[5]

The skewed anisotropic transform can be applied in any two directions with rational slopes. Instead of taking the two directions along any two random  integer lines, concept of integer lattices is used here and this will avoid directional interaction.A full-rank integer lattice A consists of the points obtained as linear combinations of two linearly independent vectors, where both the components of the vectors and the coefficients are integers. The lattice A can be represented by a non-unique generator matrix [5],

$$M_A = \begin{bmatrix} a1 & b1 \\ a2 & b2 \end{bmatrix} = \begin{bmatrix} d1 \\ d2 \end{bmatrix} \qquad (5)$$

The integer lattice Λ  is a sub lattice  of  discrete space Z which can be partitioned into det(**M**Λ) cosets of the lattice Λ, where each coset is determined by the shift vector **s**$k$, for $k$ = 0, 1, . . . ,  / det(**M**Λ)/ −1. Therefore, the lattice Λ   with the corresponding generator matrix by (5), partitions each digital line L(r1 = b1/a1, n) into co-lines. A co-line is simply the intersection between a coset and a digital line. Similarly, the digital line L(r 2 = b2/a2, n) is also partitioned into the corresponding co-lines (Fig. 3).   Notice that both filtering and sub sampling are applied in each of the cosets separately. Furthermore, each filtering operation is purely 1-D.  For constructing the Directionlets the discrete space containing the image is partitioned into integer lattices, where the 1-D filtering is performed along co-lines across the lattice. For a lattice Λ, the skewed transforms are applied along co-lines in the transform and alignment directions of the lattice Λ. The basis functions of the S-AWT are called directionlets, since they are anisotropic and have a specific direction [5].

## 2.3. Training set generation

To generate a training set, a collection of high-resolution images and their low resolution images are used. Low resolution images are formed by averaging the intensities of 2x2 pixels corresponding to the same scene in the high resolution image where 2 is the decimation factor using the equation (2). That is observed low resolution image is super resolved to its next octave. Directional information more than standard (horizontal and vertical) directions are extracted using directionlet transform. For the case of images, the information vary over space .Thus directionality can be considered as a local feature, defined in a small neighbourhood. Therefore to extract directional variations of an image it has to be analysed locally. For this the high resolution and low resolution images are subdivided into patches of size 8x8 and 4x4 respectively in raster scan order .The super-resolution algorithm operates under the assumption

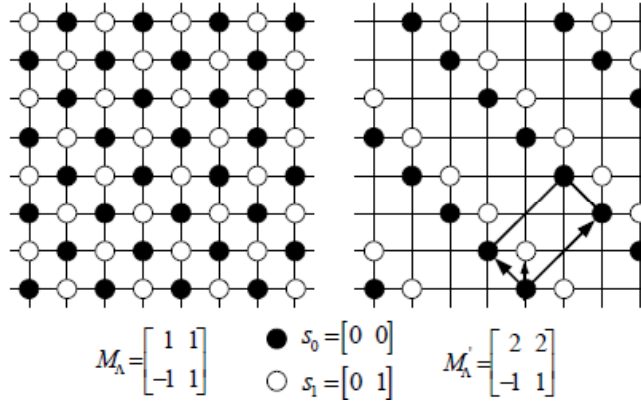

$M_\Lambda = \begin{bmatrix} 1 & 1 \\ -1 & 1 \end{bmatrix}$    ● $s_0 = \begin{bmatrix} 0 & 0 \end{bmatrix}$  ○ $s_1 = \begin{bmatrix} 0 & 1 \end{bmatrix}$    $M_\Lambda^{'} = \begin{bmatrix} 2 & 2 \\ -1 & 1 \end{bmatrix}$

Figure 4 Lattice partitions the cubic lattice into cosets along 45° and -45°, then the sub sampling are applied separately in two cosets [9].

that the predictive relationship between low and high-resolution images is independent of local image contrast. Because of this, patch pairs of high and low resolution images are contrast normalised by the energy of the low-frequency patch. This energy is the average absolute value of the low-resolution patch:

$$energy = 0.01 + \sum_i \sqrt{y_{i^{\wedge}2}} \qquad (6)$$

where $y_i$ is the value of the pixel number i , in the low-resolution patch. The 0.01 constant is added to prevent from dividing by zero. The best pair of directions for each patch is chosen from five sets of directions [(0, 90), (0, 45), (0,-45), (90, 45), (90,-45). The assigned pairs of transform directions to each patch across the low resolution image domain form a directional map of that image and its high resolution image. For this directionlet transform is applied in each patch along these five set directions. The best pair of directions $d_n$ is chosen for each patch indexed by n as

$$d_n = \arg\min \sum_{n,i} |W_{n,i}|^{\wedge}2 \qquad (7)$$

where the wavelet coefficients $W_{n,i}$ are produced by applying directionlets to the $n^{th}$ patch along the pair $d_n$ of directions. The directional map determined by the set $\{d^*\}$ minimizes the energy in the high-pass sub bands and provides the best matching between transform and locally dominant directions across segments [5]. Instead of using the critically sampled version of directionlets (filtering operations followed by sub sampling), here an oversampled version of directionlets (without sub sampling) is used for making the extraction of edge information more robust.Directionlets are then applied in each patch of the low resolution and high resolution image along the selected best pair of directions from the set D = f (0, 90), (0, 45±), (90, 45±) using the Daub4 basis functions. In the training set, the coefficients of six sub bands HL, HH, VL, VH, DL, and DH of low and high resolution image patches are stored in lexicographical order. By grouping the patch coefficients according to the direction, the searching time can be considerably reduced.

## 2.4. Prediction

The super-resolution algorithm predicts the next octave up of an input image —that is, the frequencies missing from an image zoomed with cubic spline interpolation. The interpolation of an image does not suffer any degradation, if there are no edges or contours. However, if it contains edges, they get blurred during the up sampling process. The idea used here is to learn the High Resolution (HR) representation mapping of a Low Resolution (LR) edge from the training data set during up sampling .Directionlet transform  can be used here for extracting directional features of different high resolution images and these directional features can be used here for super resolving an input low resolution image. For this a training set is generated using the directionlet transform coefficients of patches of high resolution images and their corresponding low resolution images. The directionlet coefficients of patches of  low resolution image to be super resolved are compared with those of low resolution images in the training set and if a matching is found there the directionlet coefficients of the corresponding high resolution image patch  are used as the directionlet coefficients of the high resolution image patch of the input low resolution image.

## 2.5 Learning directionlet coefficients

The given contrast normalized low resolution image to be super resolved is also sub divided into patches of size 4*4 and each patch is  decomposed into eight directional sub bands using directionlets. The best direction pair for each patch is found out as in the case of training set images. The idea used here is that the cubic spline interpolation of the input image makes the approximation A (AL and AH). The directional coefficients of other six bands HL, HH, VL, VH, DL, and DH corresponding to the finest level are learned from the training set. Here minimum absolute difference (MAD) criterion is used to select the directionlet coefficients.

The absolute difference between directionlet coefficients of the patches of the input image and those of several low resolution images in the training set is taken. The concept used here is that cubic spline interpolated image contains only low and mid frequencies and lacks high frequencies. Usually high resolution image contains low, mid and high frequencies. To super resolve a low resolution image patches of the cubic spline interpolated low resolution image is used as the approximation or low frequency bands (low and mid). The high frequency bands HL, HH, VL, VH, DL, DH are learned from the training set .These learned high frequency bands and  cubic spline interpolated low resolution input image patch are used to reconstruct the high resolution equivalent of the low resolution input. In effect, best matching 8*8 coefficients are obtained from the training data for a given 4*4 patch in the low resolution image. At the end the contrast normalization is undone.

# 3 IMPLEMENTATION AND RESULTS

Experiments are performed for various types of images with one training set with ten good quality images with different information content. The training set is not specific to the class of objects to be super resolved. All the training set images are down loaded from internet. Low resolution image of any size can be super resolved with this training set. This method is effective in the case of aliased images also.

The directionlet method has been compared with the cubic spline interpolation method and two wavelet transform based methods. The first wavelet method is proposed    by Jiji C.V and M .V Joshi (2000) .This method was also based on the assumption that images could be decomposed into low and high frequency bands using wavelet transform. To super resolve a low resolution image, that input low resolution image was considered as approximation or low frequency band of the unknown high resolution image. For obtaining high frequency band, a training set containing high resolution images was used here. Two level wavelet decomposition of the   low resolution image and three level decomposition of the training set images were taken and used as the training set. The absolute difference between the wavelet coefficients in the low resolution image and corresponding coefficients for each of training set images was taken. Parent child relationship was considered while comparing. If a matching was found here, the corresponding high resolution coefficients were taken as the unknown high frequency coefficients. The high resolution image was obtained    by taking the inverse wavelet transform of the approximation and learned high frequency bands. In order to bring in a spatial coherence during the high resolution reconstruction, a smoothness constraint is needed.  This method without smoothing was implemented here.

  In the second method instead of applying wavelet transform in the full image,  the high resolution images and their low resolution images are divided into small blocks of 8x8 and 4x4 respectively and wavelet transform was applied to these patches.  The wavelet coefficients of these patches were stored in the training set. The low resolution image to be super resolved was also divided into patches and wavelet coefficients of these patches was compared with those patches in the training set .The patch with minimum absolute difference was found out and the wavelet coefficients of its high resolution patch was used as the high resolution coefficients of the input low resolution patch. Here also the low resolution patch was considered as the approximation. The inverse wavelet transform gave high resolution patch of the low resolution input patch. The process was repeated for all the patches in the low resolution image in raster scan order to obtain the unknown high resolution image It is found that the first   wavelet based method has some disadvantages. One problem is that it needs regularization to bring spatial coherence. Another problem is that it is highly resolution dependent. For example if one wants to super resolve an image of  size MxM, the training set images must be of size 2Mx2M. If one wants to super resolve an image of any other size NxN none of the existing database

### TABLE 1.COMPARISON OF MEAN SQUARE ERROR

| Method | Butterfly | Barbara | star |
|---|---|---|---|
| Cubic     spline method | 0.0264 | 0.0447 | 0.0285 |
| Wavelet Method 2 | 0.0260 | 0.0440 | 0.0282 |
| Method   using directionlet. | 0.0239 | 0.0417 | 0.0269 |

could be used for training. This problem is solved in the second wavelet method by using block based approach. But this method also fails to eliminate aliasing.

An objective comparison of the super resolved images is done by calculating the mean squared error using equation (8).The MSE values are shown in Table 1.

$$MSE = \frac{\sum_{i,j} [z(i,j) - z'(i,j)]^2}{\sum_{i,j} z(i,j)^2} \qquad (8)$$

where z(i ,j) and z'(i, j) is the (i ,j)[th] pixel intensity of original image and reconstructed image respectively. These values also show that images obtained using directionlet are better than the ones with other standard methods.

To obtain a subjective comparison of the reconstructed images ,the super resolved images are given below. Results obtained on a low resolution image where aliasing is high, are shown here. Fig 5(a) shows such a low resolution image. Figure 5(c) and 5(d) are high resolution images obtained by cubic spline and wavelet method 2. In these images there are cross lines in the stripes of the scarf which makes them aliased. Fig 5(e) shows the super resolved image obtained using the proposed method. It is seen that aliasing is reduced considerably here since there are less cross lines and directions of stripes are almost same as original image. The visual quality of this image can be improved by adding more suitable images in the training set.

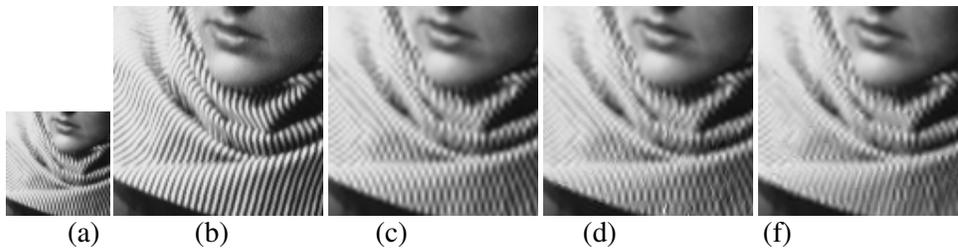

|   (a)   |   (b)   |   (c)   |   (d)   |   (f)   |

Figure 5 (a) low resolution image (b) original image (c) cubic spline interpolated image (d) super resolved image using wavelet based method 2 (e) super resolved image using proposed method

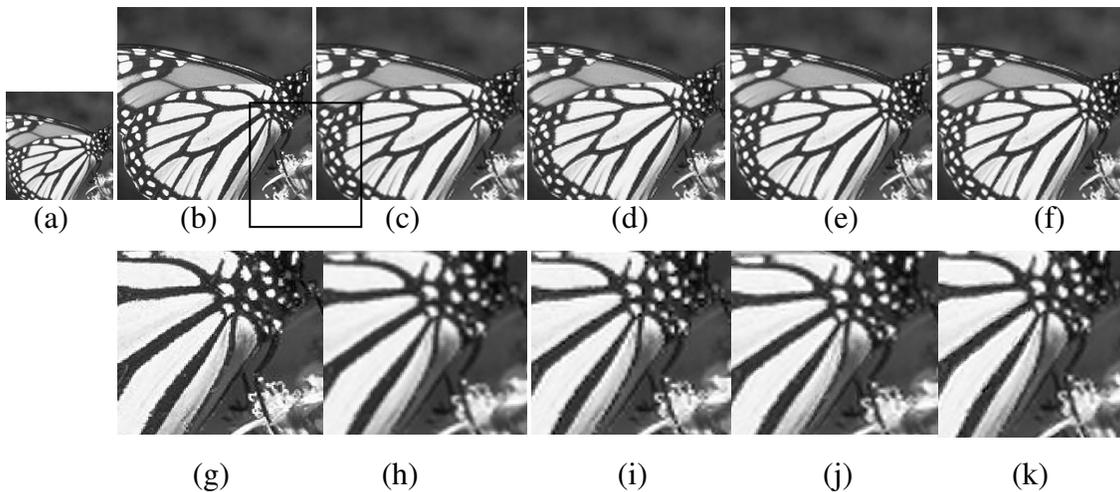

|   (a)   |   (b)   |   (c)   |   (d)   |   (e)   |   (f)   |

|   (g)   |   (h)   |   (i)   |   (j)   |   (k)   |

Figure 6 (a) low resolution image (b)original high resolution image (c) cubic spline interpolated image (d) super resolved image using wavelet based method 1(without smoothing) (e) super resolved image using wavelet based method 2 (f) super resolved image using proposed method.(g),(h),(i),(j) are zoomed portions of marked portion of (b),(c),(d),(e),(f) respectively.

The results of experiments done on a natural image are shown in Figure 6. Figure 6(a) shows the low resolution image of the high resolution image 6(b). Figure 6(c), 6(d), 6(e), 6(f) show the results obtained using standard cubic spline interpolation, first    and second wavelet methods and directionlet methods respectively. Figures 6(h), 6(i), 6(j), 6(k)   are the zoomed portions of the marked portion in Fig 6(c), 6(d), 6(e), 6(f). The ringing effects on the sharp dark edges are visible in Fig 6(h), 6(i), 6(j) whereas it is eliminated in Fig 6(k).

## 3. CONCLUSIONS

Here a directionally adaptive single image super resolution method is presented. This method uses skewed anisotropic wavelet transform called directionlets to super resolve an image. To make the edge extraction more precise, oversampled version of directionlet is used here. Images are analysed blockwise since directional information vary locally in images. This method suitably adjusts the transform directions to dominant directions of the each segment (block) of image and captures the feature variations.  Directionlet coefficients of higher sub bands at finer scale of an input low resolution image are learned from the training set of high resolution images and its high resolution image is reconstructed by taking the inverse directionlet transform. Daub4 wavelet basis was used here .Other wavelet basis like biorthogonal 9-7 can be tried instead to find the suitable basis function. Visual quality of super resolved images using directionlet is much better than that of cubic spline interpolated images and also super resolved images using discrete wavelet transform. The mean square values also show that this new method outperforms the other standard methods. The quality of the super resolved image can be improved by increasing the number of suitably selected training set images. Large size of the training set and computational burden are the two drawbacks of the proposed method .Future works concentrate on these problems. Searching time can be considerably reduced, if the size of the training set is reduced. Computational burden can be reduced by using lifting scheme which is proved to be effective in the case of wavelets.

## Authors

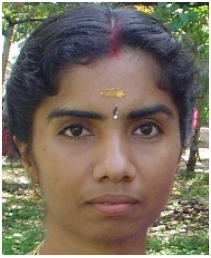

**Reji A.P** received M Sc(Electronics) and M.Tech Degree in Digital Electronics from Cochin University of science and technology ,Cochin-22,India .She has been working as Lecturer in Dept. Of Electronics in N.S.S College, Rajakumari, Kerala, India. She is doing Ph.D in Department of Electronics, Cochin University of science and Technology, India. Her research interest include Image Processing, Computer Vision ,Signal processing**.**

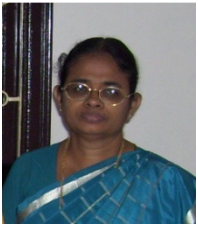

**Dr.Tessamma Thomas** received her M.Tech. and Ph.D from Cochin University of Science and Technology, Cochin-22, India. At present she is working as Professor in the Department of Electronics, Cochin University of Science and Technology. She has to her credit more than 80 research papers, in various research fields, published in International and National journals and conferences. Her areas of interest include digital signal / image processing, bio medical image processing, super resolution, content based image retrieval, genomic signal processing, etc.